\documentclass[twoside]{article}

\usepackage[accepted]{aistats2024}

\usepackage[utf8]{inputenc} % allow utf-8 input
\usepackage[T1]{fontenc}    % use 8-bit T1 fonts

\usepackage[usenames,dvipsnames]{xcolor}
\usepackage[colorlinks,linktoc=all]{hyperref}       % hyperlinks
\usepackage{url}            % simple URL typesetting
\hypersetup{citecolor=MidnightBlue}
\hypersetup{urlcolor=MidnightBlue}
\hypersetup{linkcolor=black}

\usepackage{url}            % simple URL typesetting
\usepackage{booktabs}       % professional-quality tables
\usepackage{amsfonts}       % blackboard math symbols
\usepackage{nicefrac}       % compact symbols for 1/2, etc.
\usepackage{microtype}      % microtypography
\usepackage{xcolor}         % colors

\usepackage{natbib}			% to suppress citet error highlighting
\usepackage[textsize=tiny]{todonotes}
\usepackage{graphicx}
\usepackage{amsfonts}       % blackboard math symbols
\usepackage{amsmath}
\usepackage{amssymb}
\usepackage{booktabs}       % professional-quality tables
\usepackage{epsfig}
\usepackage{microtype}      % microtypography
\usepackage{subcaption}
\usepackage{threeparttable}
\usepackage[utf8]{inputenc} % allow utf-8 input
\usepackage{url}            % simple URL typesetting
\usepackage{nicefrac}       % compact symbols for 1/2, etc.
\usepackage{xcolor}         % colors
\usepackage{float}
\usepackage{listings}
\usepackage{cleveref}
\usepackage{amsopn}
\usepackage{bm} % bold symbol
\usepackage{bbm}
\usepackage{enumerate}
\usepackage[english]{babel}
\usepackage{amsthm}
\theoremstyle{definition}
\newtheorem{definition}{Definition}

\theoremstyle{proposition}
\newtheorem{proposition}{Proposition}

\DeclareMathOperator*{\argmax}{arg\,max}

\newcommand{\vct}[1]{\boldsymbol{\mathbf{#1}}} % vector
\newcommand{\X}{\vct{X}}

\newcommand{\y}{\vct{y}}

\begin{document}
\twocolumn[

\aistatstitle{Density Uncertainty Layers for Reliable Uncertainty Estimation}

\aistatsauthor{Yookoon Park \And David M. Blei}

\aistatsaddress{Department of Computer Science \\
						Columbia University \\
						New York, NY 10027, USA \\ 
						\texttt{yookoon.park@columbia.edu} \\
						\And  
						Department of Computer Science, Statistics \\
						Columbia University \\
						New York, NY 10027, USA \\
						\texttt{david.blei@columbia.edu} \\
						 }]

%\title{Density Uncertainty Layers for \\ Reliable Uncertainty Estimation}

%\author{%
%  Yookoon Park \\
%  Department of Computer Science \\
%  Columbia University \\
%  New York, NY 10027, USA \\
%  \texttt{yookoon.park@columbia.edu} \\
%  \And
%  David M. Blei \\
%  Department of Computer Science, Statistics \\
%  Columbia University \\
%  New York, NY 10027, USA \\
%  \texttt{david.blei@columbia.edu} \\
%}

%\maketitle

\begin{abstract}
Assessing the predictive uncertainty of deep neural networks is crucial for safety-related applications of deep learning. Although Bayesian deep learning offers a principled framework for estimating model uncertainty, the common approaches that approximate the parameter posterior often fail to deliver reliable estimates of predictive uncertainty. In this paper, we propose a novel criterion for reliable predictive uncertainty: a model's predictive variance should be grounded in the empirical density of the input. That is, the model should produce higher uncertainty for inputs that are improbable in the training data and lower uncertainty for inputs that are more probable.  To operationalize this criterion, we develop the \textit{density uncertainty layer}, a stochastic neural network architecture that satisfies the density uncertain criterion by design.  We study density uncertainty layers on the UCI and CIFAR-10/100 uncertainty benchmarks. Compared to existing approaches, density uncertainty layers provide more reliable uncertainty estimates and robust out-of-distribution detection performance.
\end{abstract}

\section{Introduction}

The success of deep learning models in a range of applications has spurred
significant interest in deploying the for real-world
predictions. But in high-stakes domains, such as healthcare,
finance, and autonomous systems, both the model's prediction and its
predictive uncertainty are crucial. A challenge is that
conventional neural networks lack a robust mechanism for quantifying uncertainty, and they tend to produce overconfident predictions
\citep{guo2017calibration, ovadia2019can}. This paper is about how to
produce reliable estimates of predictive uncertainty with deep neural networks. 

%In this work, we propose a novel uncertainty criterion for reliable uncertainty estimation. The proposed \textit{density uncertainty} criterion posits that a model's predictive uncertainty should be grounded in the empirical density of the input, with higher uncertainty for improbable inputs and lower for more probable ones. 
%As a motivating example, we illustrate how Bayesian linear regression inherently adheres to the criterion. 
%To operationalize the idea, we introduce an energy-based model of input and implement the density uncertainty criterion as an explicit constraint on the predictive variance such that the variance is proportional to the input's energy. 
%We subsequently develop the density uncertainty layer, a stochastic neural network layer that derives its predictive uncertainty from a Gaussian energy model of the input. By design, the density uncertainty layers satisfy the proposed uncertainty criterion and serve as a flexible building block for uncertainty-aware deep neural networks.
%Using the CIFAR-10 and CIFAR-100 classification benchmarks, we demonstrate the density uncertainty layer provides reliable uncertainty estimates and robust out-of-distribution (OOD) detection performance for deep learning. 

\begin{figure*}[t!]
	\centering
	\begin{subfigure}{0.24\textwidth}
		\centering
		\includegraphics[width=\textwidth]{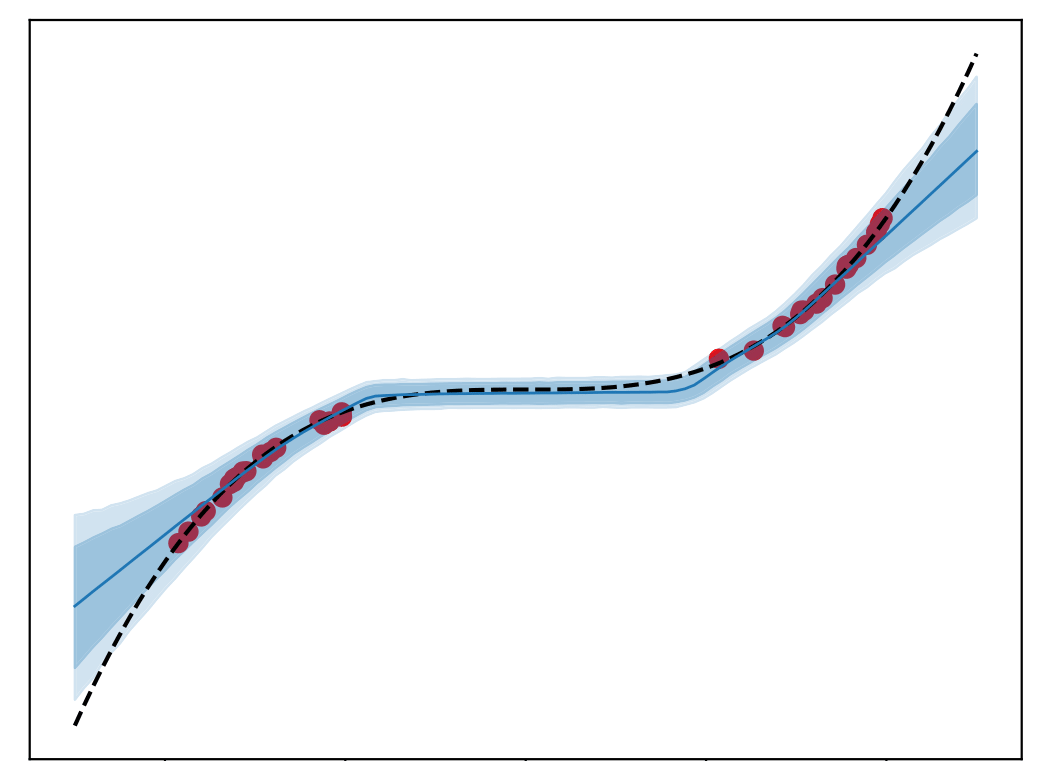}
		\caption{MFVI}
	\end{subfigure}
	\begin{subfigure}{0.24\textwidth}
		\centering
		\includegraphics[width=\textwidth]{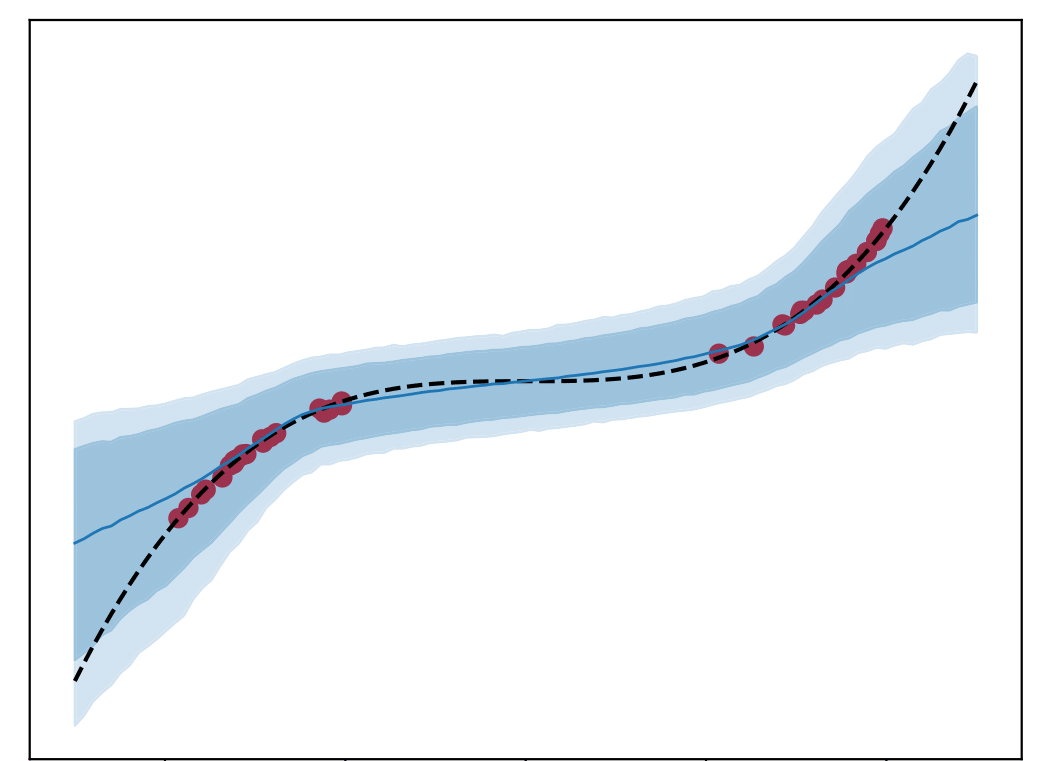}
		\caption{Variational Dropout}
	\end{subfigure}
	\begin{subfigure}{0.24\textwidth}
		\centering
		\includegraphics[width=\textwidth]{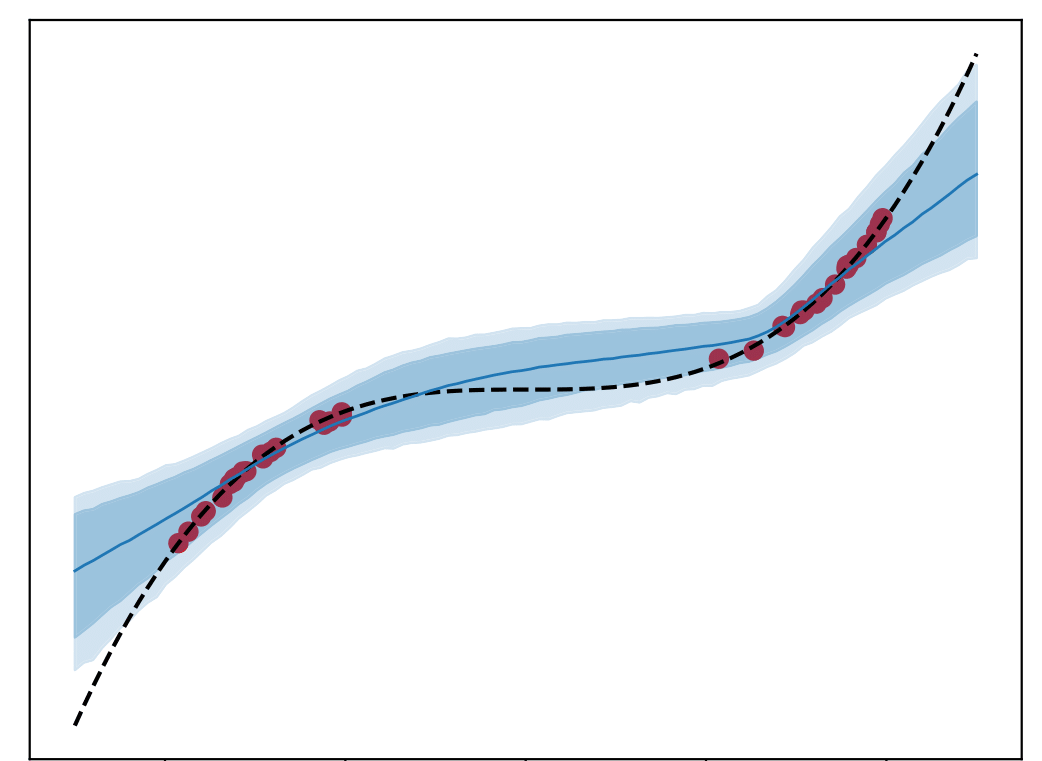}
		\caption{Rank 1 BNN}
	\end{subfigure}
	\begin{subfigure}{0.24\textwidth}
		\centering
		\includegraphics[width=\textwidth]{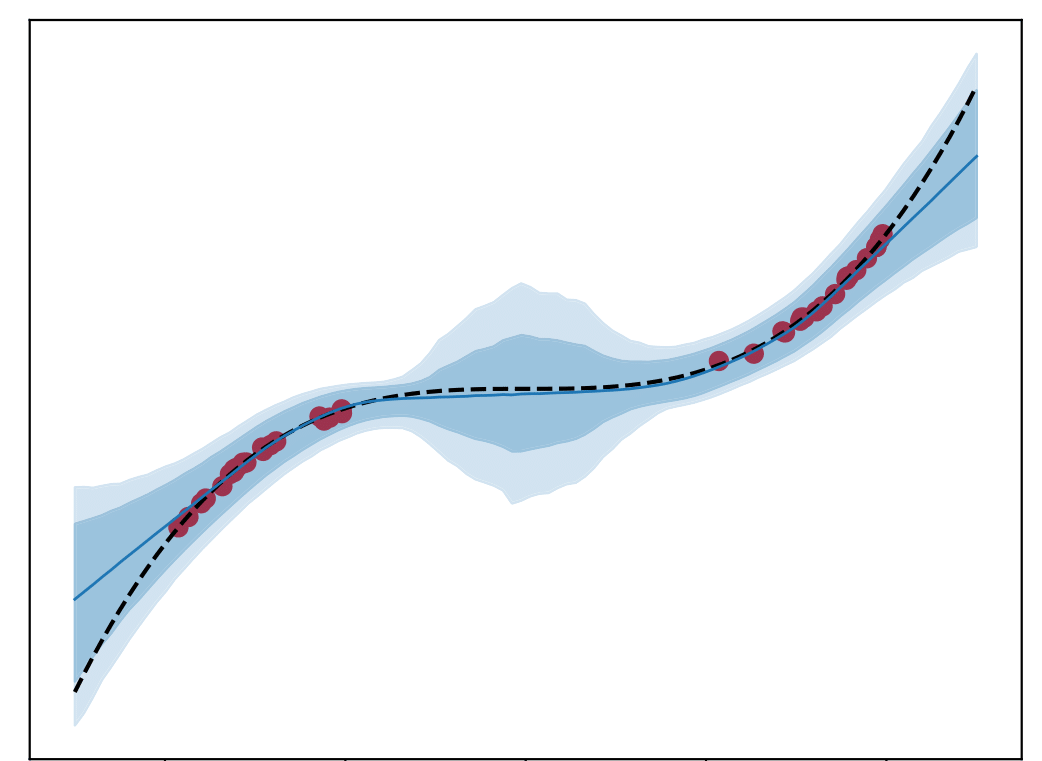}
		\caption{Density Uncertainty}
	\end{subfigure}
	\caption{Predictive uncertainty of a two-layer MLP on a toy regression problem. The red dots denote the training data and the blue shades mark the 95th and 99th percentiles of the predictive variance. All baselines other than Density Uncertainty fail to account for the in-between uncertainty in the low-density region around the origin}
	\label{fig:toy}
\end{figure*}

Why is this a problem? Bayesian deep learning offers a principled
framework for quantifying predictive uncertainty by
incorporating uncertainty about the model
parameters~\citep{graves2011practical, welling2011bayesian,
  neal2012bayesian, blundell2015weight}.  However, the common variational inference (VI) approaches 
that approximate the parameter posterior in Bayesian deep learning~\citep{graves2011practical, blundell2015weight, kingma2015variational, dusenberry2020efficient} often fall short of
providing reliable predictive uncertainty~\citep{foong2019between,
  ober2019benchmarking}. For example, 
\Cref{fig:toy}a-c demonstrate that these methods produce either collapsed or flat predictive uncertainty around the origin, even though no training data were observed in that region. 

In this work, we propose a novel criterion for reliable predictive
uncertainty and a new stochastic neural
network architecture to satisfy the criterion. The \textit{density uncertainty criterion}
posits that a model's predictive uncertainty should be grounded in the
empirical density of the input. That is, we should see higher uncertainty for
improbable inputs and lower for more probable ones.  As motivation, we
will show how Bayesian linear regression inherently adheres to
the criterion.

We then develop the \textit{density uncertainty layer}, a stochastic neural network architecture that is designed
to satisfy the density uncertainty criterion. The idea is to fit an
energy-based model of input and then satisfy the density uncertainty
criterion via a constraint on the predictive variance produced by the
approximate posterior. Density uncertainty layers serve as a flexible building block for
density-aware deep neural networks. See \Cref{fig:toy}d for how Density Uncertainty produces high predictive uncertainty in the low-density region in the input space around the origin.

We study density uncertainty layers on the UCI and 
CIFAR-10/100 uncertainty benchmarks. 
We compare the proposed method to popular
uncertainty estimation methods for neural
networks~\citep{blundell2015weight,kingma2015variational,gal2016dropout,dusenberry2020efficient}. 
We find that the proposed density-aware neural
networks provide more reliable predictive uncertainty estimates and
robust out-of-distribution (OOD) detection performance.

\textbf{Contributions.} In summary, this paper makes the following contributions.
\begin{enumerate}
\item We propose the \textit{density uncertainty criterion} for reliable predictive uncertainty, asserting that a model's predictive uncertainty should be grounded in the training density of the input.

\item We present the \textit{density uncertainty layer} whose predictive uncertainty adheres to the density uncertainty criterion and serves as a general building block for uncertainty-aware neural networks. 

\item We study the proposed density uncertainty layers on the UCI and CIFAR-10/100 uncertainty estimation benchmarks. On both benchmarks, this method
  performs better than existing approaches.
\end{enumerate}

%%% Local Variables:
%%% mode: latex
%%% TeX-master: "main"
%%% End:

\section{Density Uncertainty Layers}
\subsection{Motivation: Bayesian Linear Regression}
\label{sec:bayesian_linear_model}
We first illustrate how Bayesian uncertainty for linear regression is grounded in the empirical density of the input. 
In the next section, this observation will motivate a novel criterion for reliable predictive uncertainty.

Consider a Bayesian linear regression model with input $\X \in \mathbb{R}^{N \times D}$, target $\y \in \mathbb{R}^N$, and weight $w \in \mathbb{R}^D$:
\begin{align}
 p(w) 				&= \mathcal{N}(w| 0, \alpha^{-1} I), \label{eq:bayesian_linear} \\ 
 p(\y | \X, w)  &= \prod_{i=1}^N \mathcal{N}(y_i| w^T x_i, \beta^{-1}), \label{eq:bayesian_linear_end} 
\end{align}
The posterior distribution of the weight $w$ given the observations $\mathcal{D} = \{\X, \y\}$ is
\begin{gather}
p(w | \mathcal{D}) = \mathcal{N}(w| \mu, \Lambda^{-1}), \label{eq:bayesian_linear_posterior_start}
\\ \text{ where } \mu = \beta \Lambda^{-1} \X^T \y
\text{ and } \Lambda = \beta \X^T \X + \alpha I. \label{eq:bayesian_linear_posterior_end}
\end{gather}
The posterior predictive distribution for a test input $x_*$ is obtained by marginalizing out the weight from the posterior joint
\begin{align}
p(y_*| x_*, \mathcal{D}) 
&= \int p(y_*, w | x_*, \mathcal{D}) dw 
%\\&= \int p(y_*|x_*, w) p(w | \mathcal{D}) dw 
%\\&= \int \mathcal{N}(y_*| w^T x_*, \beta^{-1}) \mathcal{N}(w| \mu, \Lambda^{-1}) dw
\\&= \mathcal{N}(y_* | \mu^T x_*, \beta^{-1} + x_*^T \Lambda^{-1} x_*).
% \\&= \mathcal{N}(\mu^T x_*, \beta^{-1}(1 + \underbrace{x_*^T (\X^T \X + \frac{\alpha}{\beta} I)^{-1} x_*}_\text{Energy of $x_*$}))
\label{eq:bayesian_linear_posterior_predictive}
\end{align}

\Cref{eq:bayesian_linear_posterior_predictive} helps establish the connection between the Bayesian predictive uncertainty for linear regression and the empirical input density. 
Rewriting the predictive variance in \Cref{eq:bayesian_linear_posterior_predictive},
\begin{align}
\text{Var}[y_* | x_*, \mathcal{D}]  
%&= \beta^{-1} + x_*^T (\beta \X^T \X + \alpha I)^{-1} x_* \\
&= \frac{1}{\beta} + \frac{2}{\beta N} E(x_*), \label{eq:bayesian_linear_predictive_variance} \\
\text{ where } E(x_*) &= \frac{1}{2}x_*^T {\Sigma}^{-1} x_* , \label{eq:gaussian_energy}\\
{\Sigma} &= \frac{1}{N}(\X^T \X + \frac{\alpha}{\beta} I). \label{eq:bayesian_linear_covariance}
\end{align}
Here $E(x)$ is an \textit{energy function}, an unnormalized negative log density of $x$ such that
\begin{align}
p_{\text{energy}}(x) \propto \exp(-E(x)) 
%= \exp(-\frac{1}{2} x^T \Sigma^{-1} x).
\end{align}
Specifically, \Crefrange{eq:gaussian_energy}{eq:bayesian_linear_covariance} describe an instance of Gaussian energy model $\mathcal{N}(0, {\Sigma})$,
%This energy model assumes the data come from a zero-mean normal distribution $x \sim \mathcal{N}(0, \Sigma)$ with covariance $\Sigma$. 
where ${\Sigma}$ is the MAP estimate of the input covariance matrix given the observations $\X$.
This shows that Bayesian linear model derives its predictive uncertainty from the Gaussian energy model of the input that is fitted to the training data. Consequently, predictive uncertainty will be high for test inputs that are improbable in the training density and low for those that are more probable, providing intuitive and reliable predictive uncertainty. 
In the following sections, we build upon this principle.
%to develop a novel criterion for reliable uncertainty estimation. 

\subsection{Variational Inference for BNNs}
A Bayesian Neural Network (BNN) $\Omega(x; \theta)$ treats its parameters $\theta$ of neural networks as random variables from a prior $p(\theta)$. 
Combined with a likelihood function $p(y|\Omega(x; \theta))$ and the observations $\mathcal{D} = \{(x_1, y_1), ..., (x_N, y_N)\}$, the posterior distribution of the parameters is defined by the Bayes' rule:
\begin{align}
p(\theta|\mathcal{D})  \propto p(\theta) \prod_{i=1}^N p(y_i|\Omega(x_i; \theta))
\end{align}
However, the posterior is highly complex and intractable in BNNs. Variational inference (VI) circumvents this problem by optimizing a tractable approximation to the posterior. For example, the approximate distribution $q(\theta)$ commonly takes the form of fully-factorized Gaussians whose parameters are denoted by $\phi$. VI fits the approximate posterior by maximizing the evidence lower-bound (ELBO) \citep{blundell2015weight}:
\begin{align}
\mathcal{L}_{\phi}
&= \mathbb{E}_{q(\theta)}\Big[\sum_{i=1}^N \log p(y_i | \Omega(x_i; \theta))\Big] - D(q(\theta)\| p(\theta)), \label{eq:elbo} \nonumber
\end{align}
where $D$ is the Kullback-Leibler divergence. At test time, the posterior predictive distribution is approximated with $M$ Monte Carlo samples from $q(\theta)$:
\begin{align}
p(y_*|x_*, \mathcal{D}) \approx \frac{1}{M} \sum_{m=1}^M p(y_*|\Omega(x_*; \theta_m)), 
\end{align}

\subsection{The Density Uncertainty Criterion}
\label{sec:density_uncertainty_criterion}

We have shown that Bayesian predictive uncertainty has desirable properties for a linear model. 
%But consider using variational inference (VI) for approximating the intractable posterior of a nonlinear neural network model. In this setting, the corresponding approximate posterior predictive distribution often fails to provide reliable uncertainty estimates \citep{foong2019pathologies, ober2019benchmarking}. 
%This issue is also evident in the toy nonlinear regression problem in \Cref{fig:toy} where the VI baselines fail to accurately capture the density of the data in their predictive uncertainty. 
However, VI-based BNNs often fails to provide reliable uncertainty estimates \citep{foong2019pathologies, ober2019benchmarking} as demonstrated in \Cref{fig:toy} where the VI baselines fail to accurately capture the density of the data in their predictive uncertainty.
%But for a nonlinear neural network model with intractable posterior, the corresponding approximate posterior predictive distribution often fails to provide reliable uncertainty estimates \citep{foong2019pathologies, ober2019benchmarking}. 

To address this, we propose a novel uncertainty criterion that asserts model's predictive uncertainty should be grounded in the training density of the input. To formalize the concept, we first introduce an auxiliary energy-based generative model of the input:
\begin{align}
p_{\text{energy}}(x; \omega) \propto \exp(-E(x; \omega)),
\end{align}
where the generative parameter $\omega$ is fitted to the training data $x_1, ..., x_N$. We omit $\omega$ for simplicity hereafter. 

We now establish the \textit{density uncertainty criterion}.
\begin{definition}[\textbf{Density uncertainty criterion}]
	For a parameterized model $f(x; \theta)$ with a distribution on the parameters $q(\theta)$ and an non-negative energy model $E(x) \ge 0$ fitted to the training data, the model $f(x; \theta)$ adheres to the density uncertainty criterion if
	 \begin{align}
	 \text{Var}_{q(\theta)}[f(x; \theta)] \propto E(x) \text{ for all } x \in \mathcal{X}. \label{eq:uncertainty_criterion}
	 \end{align}
\end{definition}
The criterion posits that the predictive uncertainty should be proportional to the energy of the input,
producing high predictive uncertainty for inputs that are improbable in the training distribution and low uncertainty for inputs that are more probable. 
As illustrated in \Cref{sec:bayesian_linear_model}, Bayesian linear regression inherently satisfies the density uncertainty criterion of \Cref{eq:uncertainty_criterion}. 

%How do we incorporate the density uncertainty criterion when fitting a model $f(x, \theta)$?
%In Bayesian deep learning, common VI approaches fit $q(\theta)$ by maximizing the evidence lower-bound (ELBO) \citep{blundell2015weight}. 
%Building on this objective, we include the density uncertainty criterion as a constraint during optimization:

%It is left to integrate the density criterion into the model optimization. We choose the evidence lower-bound (ELBO) objective commonly used in VI, yielding a constrained optimization objective

%In Bayesian deep learning, common VI approaches fit $q(\theta)$ by maximizing the evidence lower-bound (ELBO) \citep{blundell2015weight}. 
Incorporating the density uncertainty criterion into the ELBO objective yields
\begin{gather}
\argmax_{q \in \mathcal{Q}} \mathbb{E}_{q(\theta)} \Big[\sum_{i=1}^N \log p(y_i | f(x_i; \theta))\Big] - D(q(\theta) \| p(\theta)) \nonumber \\
\text{ s.t. } \text{Var}_{q(\theta)}[f(x; \theta)] \propto E(x) \text{ for all } x \in \mathcal{X}, \label{eq:uncertainty_objective1} 
\end{gather} 
where $D$ is the Kullback-Leibler divergence and $p$ is a prior. 
The constraint in \Cref{eq:uncertainty_objective1} ensures the predictive uncertainty adheres to the density uncertainty criterion, so that its predictive uncertainty is consistently derived from the training density of the input, yielding reliable uncertainty estimates. 
%This distinction highlights a paradigm shift for uncertainty quantification: instead of dealing with the abstruse parameter uncertainty, we directly regulate the predictive uncertainty of the model, adhering to the density uncertainty criterion. 

\textbf{Example: Bayesian linear regression} Revisiting the Bayesian linear regression example (\Cref{sec:bayesian_linear_model}), the family of parameter distribution $q(w)$ that satisfies the density uncertainty criterion is 
\begin{align}
q(w) &= \mathcal{N}(\mu, \gamma {\Sigma}^{-1}),  \label{eq:example_linear}  \\
\text{ where } {\Sigma} &= \frac{1}{N} (\X^T \X + \frac{\alpha}{\beta} I), 
\end{align}
where $\gamma$ is a scaling scalar and $\mu$ is a trainable parameter. 
The posterior precision of the weight is now tied to the training covariance estimate of the input. This results in the predictive variance of
\begin{align}
\text{Var}_{q(w)}[f(x_*; w)] \propto E(x_*) =  \frac{1}{2} x_* \widehat{\Sigma}^{-1} x_* ,
\end{align}
satisfying the density uncertainty criterion for a Gaussian energy model $E$.
%where the Gaussian energy model $E$ is fitted to the training data. Thus, the constraint on the precision matrix acts as a form of posterior regularization that encodes our density uncertainty criterion. 
Notably, the constrained parameter posterior (\Cref{eq:example_linear}) recovers the true posterior in Bayesian linear regression (\Cref{eq:bayesian_linear_posterior_start}).

\textbf{Reparametrized objective} More generally, we introduce a reparametrized version of the objective in \Cref{eq:uncertainty_objective1}. 
Consider a stochastic function $f(x, \epsilon; \phi)$ but now with a deterministic parameter $\phi$, exogenous noise variable $\epsilon$, and a noise distribution $q(\epsilon)$:
\begin{gather}
\argmax_{\phi} \mathbb{E}_{q(\epsilon)} \Big[\sum_{i=1}^N \log p(y_i | f(x_i, \epsilon; \phi))\Big] - D(q(\epsilon) \| p(\epsilon)) \nonumber \\
\text{ s.t. } \text{Var}_{q(\epsilon)}[f(x, \epsilon; \phi)] \propto E(x) \text{ for all } x \in \mathcal{X}, 
\end{gather}
A broad class of $q(\theta)$ such as normal distributions admits this kind of reparametrization. 
%In this form, the exogenous noise $\epsilon$ sources the stochasticity of the model and exposes the deterministic model parameter $\phi$. 
The reparametrized objective offers more flexibility in how we incorporate noise into the model while adhering to the density uncertainty criterion. 
For example, we may choose to directly inject noise into the low-dimensional function output instead of sampling the high-dimensional parameters, thereby improving the computational efficiency and reducing the gradient variance (e.g. \citet{kingma2015variational}). 
Therefore, we default to this form in the remainder of the paper. 

\subsection{The Density Uncertainty Layer}
We now select an appropriate energy model $E(x)$ and the structure of a stochastic function $\Omega(x, \epsilon; \phi)$ for deep neural networks. 
In this work, we consider a popular form of residual network architecture %\footnote{Note that this is not exactly the same as the ResNet architecture we use in the experiments}
\begin{align}
a_1 &= f_1(x, \epsilon_1; \phi_1), \\
h_\ell &= \pi(a_\ell), \\
a_{\ell+1} &= a_\ell + f_\ell(h_\ell, \epsilon_\ell; \phi_\ell), \text{ for } l = 1 , ..., L
\end{align}
where $f_l$ is a stochastic linear layer with deterministic parameter $\phi_\ell$ and exogenous noise variable $\epsilon_\ell$. $\pi$ is a nonlinear activation function such as ReLU. 
%Instead of enforcing the uncertainty criterion (\Cref{eq:uncertainty_criterion}) on the neural network $\Omega$ directly, we propose to make the individual linear layers $f_1, f_2, ..., f_{L}$ adhere to the density uncertainty criterion layer-wise. 

\Cref{sec:bayesian_linear_model} shows that Bayesian linear regression derives its predictive uncertainty from a Gaussian energy model.
Motivated by this insight, we pair each linear layer $f_\ell$ with a Gaussian energy model $E_\ell$ and make the individual linear layers $f_1, f_2, ..., f_{L}$ adhere to the density uncertainty criterion layer-wise:
\begin{gather}
\argmax_{\phi} \mathbb{E}_{q} \Big[\sum_{i=1}^N \log p(y_i | \Omega(x_i, \vct{\epsilon}; \vct{\phi}))\Big]  - D(q(\vct{\epsilon}) \| p(\vct{\epsilon})) \nonumber \\
\text{ s.t. } \text{Var}_{q(\epsilon_\ell)}[f_\ell^j(h_{\ell}, \epsilon_\ell; \phi_\ell)] \propto E_\ell(h_{\ell})  \label{eq:objective_layerwise1}  \\
\text{ for all } h_{\ell} \in \mathcal{H}, \, j=1, ..., D, \text{ and } \ell=1, ..., L,  
\end{gather}
where $j$ indexes the hidden units and $D$ is the width of the layers. 
$\vct{\phi}$ encompasses the deterministic weights of the neural network and the posterior parameters of the noise distributions, and $\vct{\epsilon}$ involves all noise variables. 

This design is based on several key observations:
%First, a Gaussian energy model provides a smooth uncertainty landscape while capturing correlations among input dimensions. 
First, the Gaussian energy facilitates efficient training and energy evaluation. 
Second, making the individual layers stochastic fosters functional diversity and stochastic regularization effect. 
Third, the complexity of the uncertainty landscape grows naturally with the model complexity with size of in the neural network.
Lastly, a relaxed version of the density uncertainty criterion holds at the network level which we describe next. 

\begin{proposition}[\textbf{Density Uncertainty Criterion for Residual Networks}]
	Define the total energy $E(x, h_1, ..., h_\ell) = \sum_{\ell=1}^L E_\ell(h_\ell)$.
	Assume the followings:
	\begin{enumerate}
		\item There exist $\alpha, \beta \in \mathbb{R}_{+}$ such that for all $j, \ell$
		\begin{align}
		\alpha E_\ell(h_{\ell}) \le \text{Var}_{q(\epsilon_\ell)}[f_\ell^j(h_{\ell}, \epsilon_\ell; \phi_\ell)] \le \beta E_\ell(h_{\ell}) \nonumber 
		\end{align}
		\item There exists $M \in \mathbb{R}_{+}$ such that $\|w_\ell^j\|_2^2 \le M$ for all $j, \ell$,
		where $w_\ell^j$ is the deterministic weight for $j$-th hidden unit at layer $l$. 
		\item $\phi$ is a 1-Lipschitz activation fucntion (e.g. ReLU). 
	\end{enumerate} 

Then the dimension-wise variance of the network output $a_{L+1}$ is bounded from below by the expected total energy:
\begin{align}
\text{Var}_{q(\theta)}[a_{L+1}^j | x] \ge C \, \mathbb{E}_{q(\theta)}[E(x, h_1, ..., h_\ell)]
\end{align}
for some constant $C$. Proof is in the appendix. 
\end{proposition}	

We now introduce the \textit{density uncertainty layer}, a stochastic architecture that by design satisfies the density uncertainty criterion layer-wise (\Cref{eq:objective_layerwise1}):
\begin{gather}
f_\ell^j(h_{\ell}, \epsilon_\ell; w_\ell^j) = w_\ell^j \cdot h_{\ell} + \epsilon_\ell^j \sqrt{E(h_{\ell})} + \eta_\ell^j,  \nonumber \\
\text{ where } E_\ell(h_{\ell}) = \frac{1}{2} h_{\ell}^T \Sigma_{\ell}^{-1} h_{\ell},  \\
q(\epsilon_\ell^j) = \mathcal{N}(0, \gamma_\ell^j), \, q(\eta_\ell^j) = \mathcal{N}(0, \beta_\ell^j), \label{eq:stochastic_linear_layer1}
\end{gather}
%for $j=1, ..., M_\ell$. 
where $\cdot$ denotes dot product, $w_\ell^j$ is the deterministic weight for the $j$-th hidden unit at layer $\ell$, %and $M_\ell$ is the number of hidden units in the layer. 
and $E_\ell(h_{\ell})$ is the Gaussian energy model with the covariance parameter $\Sigma_{\ell}$.
%The layer has two noise components: $\epsilon_\ell^j$ that scales with the energy and $\eta_\ell^j$ that is independent of the energy. 
The layer has two noise components $\epsilon_\ell^j$ and $\eta_\ell^j$, and the predictive variance of
\begin{align}
\text{Var}_{q(\epsilon_\ell)}[f_\ell^j(h_{\ell}, \epsilon_\ell; \phi_\ell)] = \gamma_\ell^j E_\ell(h_{\ell}) + \beta_\ell^j,
\end{align}
which satisfies the density uncertainty criterion as the bias term $\beta_\ell^j$ can be always absorbed into the energy function.  
The posterior noise variance parameters $\gamma_\ell^j, \beta_\ell^j$ are optimized using the ELBO, granting the model the flexibility to adjust the noise scale for individual hidden units in the layer. 
While we assume Gaussian noise for simplicity, we may potentially incorporate other noise distributions such as heavy-tailed ones \citep{dusenberry2020efficient} for enhanced robustness. 

%\subsection{Implementation}
%We now discuss the implementation details for density uncertainty layers.
 
\paragraph{Gaussian energy} 
We adopt the LDL parametrization of the precision matrix $\Sigma_{\ell}^{-1} = L_\ell D_\ell L_\ell^T$ for the layer-wise Gaussian energy models, where $L_\ell$ is a lower unit triangular matrix, and $D_\ell$ is a non-negative diagonal matrix. This admits efficient energy evaluation 
\begin{align}
E_\ell(h_{\ell}) = \frac{1}{2} h_{\ell}^T \Sigma_{\ell}^{-1} h_{\ell} = \frac{1}{2} \| D^{\frac{1}{2}}_{\ell} L_{\ell}^T h_{\ell} \|_2^2, \nonumber 
\end{align}
without computing the inverse of the covariance matrix and simplifies the log determinant of the covariance matrix as $\log |\Sigma_{\ell}| = -\sum_j \log D_{\ell}^{jj}$.
For convolutional architectures, we replace the matrix-vector product $L_{\ell}^T h_{\ell}$ with a masked convolution \citep{van2016conditional} to build a convolutional energy model. 
%This is motivated by the observation that the lower-triangular elements of $L_\ell$ can be interpreted as a linear autoregressive model of the input \citep{pourahmadi1999joint}. Then the energy can be computed as a squared sum of the residuals of the autoregressive model, weighted by $D_{\ell}^{-1}$.  
The main computational overhead of density uncertainty layers comes from the matrix-vector product $L_{\ell}^T h_{\ell}$ with $O(D^2)$ complexity, same as regular linear layers. 
%, which is roughly equivant to that of a deterministic neural network layer. 

\paragraph{Mixture of rank-1 Gaussians} When the computational overhead is a major consideration, we propose to use a mixture of rank-1 Gaussians as an alternative layer-wise energy model. 
More specifically, the rank-1 Gaussian has the covariance matrix of the form:
\begin{align}
\Sigma = v v^T + D,
\end{align}
where $v$ is the rank-1 factor and $D$ is a non-negative diagonal matrix. Using the rank-1 construction, the inverse can be computed as
\begin{align}
\Sigma^{-1} = D^{-1} - \frac{D^{-1} v v^T D^{-1}}{1 + v^T D^{-1} v},
\end{align}
and the log determinant is also simplified as
\begin{align}
\log |\Sigma| = \log (1 + v^T D^{-1} v) + \sum \log D_{jj},
\end{align}
reducing the computational overhead to $O(D)$. However, as the rank-1 construction may be overly restrictive in practice, we choose a $K$ mixture of rank-1 Gaussians as our layer-wise generative model, with the computational complexity of $O(KD)$. We set $K$ as a fraction of the layer's width and show that it can be as small as $\approx 1\%$ with negligible performance drop. 

\textbf{Optimization} The construction of the density uncertainty layer (\Cref{eq:stochastic_linear_layer1}) inherently satisfies the density uncertainty constraint, simplifying the constrained optimization objective (\Cref{eq:objective_layerwise1}) to
\begin{align}
\mathcal{L}_{\vct{\phi}} &= \mathbb{E}_{q(\vct{\epsilon})} \Big[\sum_{i=1}^N \log p(y_i | \Omega(x_i, \vct{\epsilon}; \vct{\phi}))\Big] + D(q(\vct{\epsilon}) \| p(\vct{\epsilon})) \nonumber
\end{align}
We assume Gaussian priors with a shared variance for the noise variables. 
Concurrently, the layer-wise energy models are optimized using the generative objective:
\begin{align}
\mathcal{L}_{\vct{\omega}} &= \sum_{i=1}^N  \sum_{\ell=1}^{L} \mathbb{E}_{q(\vct{\epsilon})}[\log p_{\text{energy}}(h_\ell; \omega_\ell)] + \log p(\omega_\ell), \label{eq:generative_objective} \nonumber 
\end{align}
where $p_{\text{energy}}(h_\ell; \omega_\ell) = \exp(-E(h_\ell))$ is the Gaussian energy distribution for the input at layer $\ell$ and $p(\omega_\ell)$ is a prior on the generative parameter. 
The two objectives are jointly optimized during training. 

\section{Related Work}

This paper contributes to Bayesian deep learning and uncertainty
estimation for deep learning.

\paragraph{Bayesian Neural Networks and Uncertainty} Bayesian Neural
Networks (BNNs) establish a principled framework for estimating the
uncertainty of neural networks by assuming their parameters are latent
variables that follow a prior distribution. Bayes' rule, combined with
a likelihood function and observations, defines the posterior
distribution of the parameters.  However, as exact posterior inference
in BNNs is intractable, the problem boils down to approximating the
parameter posterior distribution.  For example, Markov Chain Monte
Carlo (MCMC) \citep{welling2011bayesian, chen2014stochastic} simulates
samples from the posterior distribution, using Langevin
\citep{welling2011bayesian} or Hamiltonian \citep{chen2014stochastic}
dynamics. On the other hand, the Laplace approximation
\citep{ritter2018scalable} applies a second-order approximation at a
mode of the posterior distribution.

Variational inference (VI) is a popular approach that reformulates
inference as an optimization problem. It seeks the best approximating
distribution within a distribution family that minimizes a discrepancy
metric to the true posterior, such as the Kullback-Leibler (KL)
divergence.  \Citet{graves2011practical} adopt fully-factorized
Gaussian posteriors for the network's parameters, and
\citet{blundell2015weight} further incorporate the reparametrization
trick \citep{kingma14, rezende14} to obtain unbiased, low-variance
gradient estimates with automatic differentiation.
\Citet{louizos2016structured} enhance the expressiveness of the
posteriors by utilizing matrix Gaussian distributions for structured
modeling of parameter correlations within each layer.  More recently,
\citet{ritter2021sparse} propose sparse representations of matrix
Gaussian posteriors using inducing points
\citep{snelson2005sparse, titsias2009variational}.

However, VI often fails to provide reliable uncertainty
estimates for neural networks in practice \citep{foong2019between, ober2019benchmarking}.
This failure might stem from the disconnect between Bayesian
parameter uncertainty and predictive uncertainty---while Bayesian
methods focus on the posterior parameter uncertainty, the practical
interest often lies in estimating the model's predictive uncertainty. 
This gap may be further exacerbated by the the restrictive independence assumptions
\citep{trippe2018overpruning, foong2019pathologies,
  foong2020expressiveness} and the mode-seeking behavior of the evidence
lower-bound (ELBO)~\citep{bishop06}. \citet{sun2019functional} tries to
bridge the gap by performing variational inference in the function
space, albeit this requires additional approximations to the
intractable functional KL divergence.  In contrast, the density
uncertainty criterion directly imposes a constraint on
the model's predictive uncertainty, so that the predictive uncertainty is always
grounded in the training density of the input and provides reliable estimates of predictive uncertainty.

\paragraph{Uncertainty Estimation for Deep Learning} This paper proposes
a new methodology for estimating uncertainty in deep learning
models. Other alternatives for uncertainty estimation include Monte
Carlo dropout \citep{gal2016dropout}, which interprets dropout
regularization as approximate Bayesian inference and estimates
predictive uncertainty by performing Monte Carlo integration using
dropout at test time.  Variational Gaussian dropout
\cite{kingma2015variational}, motivated by a continuous approximation
to dropout \citep{wang2013fast}, applies multiplicative noise to the
preactivations. The authors show this corresponds to assuming a
degenerate parameter posterior distribution and propose a variational
inference method for adapting the dropout rates.  Recently,
\citet{dusenberry2020efficient} propose to model only the uncertainty
of rank-1 factors in the network's parameters by introducing dimension-wise
multiplicative noise to both the input and the output at each layer.

On the other hand, deep kernel learning methods (DKL) \citep{snoek2015scalable, wilson2016deep, liu2020simple, van2021feature} combine the expressive power of neural networks with the uncertainty estimation capability of Gaussian Processes (GPs) by 
using a deterministic neural network as a feature extractor and applying the GP in the resulting feature space. 
However, \citet{ober2021promises} show that such construction is prone to overfitting as it ignores any uncertainty associated with the neural network feature extractor. 
To alleviate the problem, \citet{liu2020simple, van2021feature} limit the expressive power of the neural network feature extractor using spectral normalization \citep{miyato2018spectral}
and apply approximations to the GP posterior to circumvent the cubic complexity of GP inference.
While DKL methods provide interesting alternatives for uncertainty estimation, 
in this work we primarily focus on the BNN and the related
methods \citep{blundell2015weight, kingma2015variational,
	gal2016dropout, dusenberry2020efficient} that model the uncertainty
of the neural network as a whole.  
 
%The neural linear model (NLM) \citep{snoek2015scalable,
%  ober2019benchmarking, brosse2020last, kristiadi2020being} adopts a
%two-stage approach for deep uncertainty estimation: first, a
%deterministic neural network is fitted using MAP and then its last
%layer is replaced with a Bayesian linear layer to estimate the
%uncertainty.  But while NLMs can provide efficient uncertainty estimates
%\citep{snoek2015scalable, riquelme2018deep, ober2019benchmarking,
%  zhou2019adaptive}, it's unclear whether the MAP estimate of the
%network would necessarily provide meaningful bases for linear
%uncertainty estimation methods.  For example, \citet{thakur2020learned,
%  watson2021latent} demonstrate that NLMs fail to capture in-between
%uncertainties and propose diversity-encouraging regularizers for the
%representations of the neural network.  \citet{liu2020simple} try to mitigate the problem by
%enforcing distance awareness into the network.

% While NLMs provide interesting alternatives for uncertainty estimation, 
%While DKL methods provide interesting alternatives for uncertainty estimation, 
%in this work we focus on the BNN and the related
%methods \citep{blundell2015weight, kingma2015variational,
%  gal2016dropout, dusenberry2020efficient} that model the uncertainty
%of the neural network as a whole. 

%But it is an interesting path of
%future work to consider the predictive uncertainty criterion in an
%NLM.

% db : it's not clear how this work fits into the NLM. are you saying
% we are pursuing a different approach.

%%% Local Variables:
%%% mode: latex
%%% TeX-master: "main"
%%% End:

\section{Empirical Studies}
\label{sec:experiments}

\begin{table*}[t]
	\caption{CIFAR-10 classification using the ResNet-14 architecture and 25 posterior samples. The average and the standard deviation over 3 random seeds are shown. Density Uncertainty significantly reduces the calibration error}
	\label{table:cifar10}
	\centering
	\begin{tabular}{l c c c}
		\toprule
		Method 	& Accuracy ($\uparrow$)  & ECE ($\downarrow$) & NLL ($\downarrow$) \\
		\midrule
		MFVI 						 & 91.9 $\pm$ 0.1 & 0.080 $\pm$ 0.002 & 0.291 $\pm$ 0.004 \\
		MCDropout 				& 91.0 $\pm$ 0.1 & 0.016 $\pm$ 0.001 & 0.261 $\pm$ 0.001\\
		VDropout 				  & 92.2 $\pm$ 0.0 & 0.013 $\pm$ 0.001 & 0.233 $\pm$ 0.002 \\
		Rank-1 BNN				& 92.2 $\pm$ 0.0 & 0.017 $\pm$ 0.002 & 0.231 $\pm$ 0.002 \\
		\midrule 
		Density Uncertainty	  & 92.2 $\pm$ 0.3 & \textbf{0.004 $\pm$ 0.000} & \textbf{0.226 $\pm$ 0.003} \\
		\bottomrule
	\end{tabular}
\end{table*}

\begin{table*}[t]
	\caption{CIFAR-100 classification using the ResNet-14 architecture and 25 posterior samples. The average and the standard deviation over 3 random seeds are shown. Density Uncertainty improves all three metrics}
	\label{table:cifar100}
	\centering
	\begin{tabular}{l c c c}
		\toprule
		Method 	& Accuracy ($\uparrow$)  & ECE ($\downarrow$) & NLL ($\downarrow$) \\
		\midrule
		MFVI 						 & 67.9 $\pm$ 0.4 & 0.132 $\pm$ 0.002 & 1.212 $\pm$ 0.017 \\
		MCDropout 				& 66.3 $\pm$ 0.3 & 0.066 $\pm$ 0.001 & 1.205 $\pm$ 0.005 \\
		VDropout 				  & 67.8 $\pm$ 0.1 & 0.051 $\pm$ 0.002 & 1.167 $\pm$ 0.002 \\
		Rank-1 BNN				& 68.0 $\pm$ 0.5 & 0.041 $\pm$ 0.003 & 1.132 $\pm$ 0.002 \\
		\midrule 
		Density Uncertainty	  & \textbf{68.8 $\pm$ 0.2} & \textbf{0.011 $\pm$  0.003} & \textbf{1.110 $\pm$ 0.006} \\
		\bottomrule
	\end{tabular}
\end{table*}

In this section, we empirically demonstrate that the density uncertainty layer delivers reliable predictive uncertainty estimates compared to the existing approaches. 
The empirical studies are structured as follows:
\begin{enumerate}
	\item We visualize the predictive uncertainty landscape of different uncertainty estimation methods and their failure modes, on a toy regression problem.
	\item We evaluate the uncertainty estimation performance on CIFAR-10/100 classification benchmarks \citep{krizhevsky09} using the ResNet-14 \citep{he2016deep} and the Wide ResNet-28 (WRN-28) \citep{zagoruyko2016wide} architecture. 
	\item We evaluate the out-of-distribution (OOD) detection performance on SVHN \citep{netzer2011reading} using the models trained on CIFAR-10/100.
\end{enumerate}
In the appendix, we include the results on the UCI regression benchmarks using a MLP architecture. 

We compare our method to the following popular uncertainty methods for deep learning:
\begin{enumerate}
	\item \textbf{Mean-field Variational Inference} (MFVI) \citep{blundell2015weight} assumes fully factorized normal posteriors on the neural network parameters. 
	\item \textbf{Monte Carlo Dropout} (MCDropout) \citep{gal2016dropout} views dropout as approximate Bayesian inference and applies dropout at test time in order to estimate the predictive uncertainty.
	\item \textbf{Variational Dropout} (VDropout) \citep{kingma2015variational} applies Gaussian multiplicative noise $\epsilon_\ell^j \sim \mathcal{N}(1, \alpha)$ to the output of linear layers.
%	\begin{align}
%	a_\ell^j = \epsilon_\ell^j w_\ell^j \cdot h_{\ell}.
%	\end{align}
	\item \textbf{Rank-1 BNN} \citep{dusenberry2020efficient} further extends Variational Dropout by introducing multiplicative noise for both the layer's input and output, dimension-wise. We assume Gaussian noise. 
\end{enumerate}
In addition, we include the last-layer GP methods of SNGP \citep{liu2020simple} and DUE \citep{van2021feature} for the WRN-28 experiments. 
We use the DUE authors’
implementation for SNGP and DUE.
%On all benchmarks, we find that the proposed method delivers more reliable predictive uncertainty than the baselines. 

\textbf{Experimental details for CIFAR-10/100} 
For the CIFAR-10/100 experiments, we use the standard convolutional ResNet-14 architecture \citep{he2016deep} and also the significantly larger WRN-28 architecture \citep{zagoruyko2016wide} with $2\times$ depth and $10\times$ width. 
We default to the full-rank Gaussian energy models using LDL parametrization but also include the efficient rank-1 mixture Gaussians for the larger WRN-28 experiments.  
We use the ADAM optimizer with learning rate $0.1$ with batch size of 128 except for MFVI where the learning rate is reduced to $0.01$ as higher learning rate led to divergence. 
%This is due to the high gradient variance stemming from the sampling of high-dimensional weights in MFVI. 
We train the models for 200 epochs with cosine learning rate schedule without restarts \citep{loshchilov2016sgdr}. During training, we apply random cropping and padding, and horizontal flipping data augmentations. The input pixel values are normalized using the training pixel means and standard deviations, channel-wise. 
We do not employ KL annealing or posterior tempering but initialize the posterior standard deviation to a sufficiently small value (e.g. $10^{-3}$) to stabilize training \citep{dusenberry2020efficient}. 
The weight decay is set to $0.0001$ for CIFAR-10 and $0.0002$ for CIFAR-100 experiments. For WRN-28, we use weight decay of $0.0005$, learning rate of $0.05$, and batch size of $64$.  

Model-specific hyperparameters are searched on a grid on a randomly sampled held-out set of CIFAR-10. 
The dropout rate for Monte Carlo Dropout is searched over $\{0.1, 0.2, 0.3\}$ and set to $0.1$. 
For Variational Dropout, we find that adapting the noise variance leads to under-fitting and thus fix the noise variance. The multiplicative noise variance is searched over $\{0.1, 0.25, 0.5\}$ and set to $0.1$. 
For Rank-1 BNN, the prior standard deviation for the multiplicative noise distributions is searched over $\{0.01, 0.1, 1\}$ and set to $0.1$. 
For Density Uncertainty, the prior noise standard deviation is searched over $\{0.1, 1, 10\}$ and set to $1$. 

All experiments are implemented in PyTorch \citep{paszke2019pytorch} and executed on a single Titan X GPU. 
The code is available at \url{https://github.com/yookoon/density_uncertainty_layers}

\subsection{Toy Regression}
We generate a toy regression problem in 1-D to illustrate the predictive uncertainty landscapes of different uncertainty estimation methods. We uniformly sample $x_i$ from $[-4, -2] \cup [2, 4]$ and generate the target as $y_i = x_i^3 + \epsilon$. This leaves a in-between low density region in $[-2, 2]$. We normalize the input and the target to have zero-mean and unit variance. We use a MLP with one hidden layer of width 50 as the base architecture. 

The predictive uncertainty landscapes on a toy regression problem are visualized in \Cref{fig:toy}. The baselines fail to produce reliable in-between uncertainty in the low-density region around the origin.
For example, MFVI gives collapsed in-between uncertainty while Variational Dropout and Rank 1 BNNs produce flat uncertainty in the low-density region. Theses two methods both apply input-independent noise to the layers and lack a robust mechanism for adjusting their uncertainty depending on the empirical density of the input.
In contrast, Density Uncertainty captures the density of the training data and produces higher predictive uncertainty in the low-density regions. This is because it derives its uncertainty from the energy model of the input, adhering to the density uncertainty criterion.
%Notably, its predictive uncertainty successfully reflects the bimodal nature of the input density even when the individual layer are associated with unimodal Gaussian energy models. This illustrates how the complexity of the uncertainty landscape in density uncertainty layers naturally grows with the number of layers in the neural network. 

\begin{table*}[t]
	\caption{CIFAR-10 classification using the WRN28 architecture and 25 posterior samples. The average and the standard deviation over 3 random seeds are shown. Density Uncertainty improves all three metrics}
	\label{table:cifar10_wrn28}
	\centering
	\begin{tabular}{l c c c}
		\toprule
		Method 	& Accuracy ($\uparrow$)  & ECE ($\downarrow$) & NLL ($\downarrow$) \\
		\midrule
		MFVI 						 & 89.3 $\pm$ 1.4 & 0.310 $\pm$ 0.011 & 0.668 $\pm$ 0.024 \\
		MCDropout 				& 96.0 $\pm$ 0.2 & 0.013 $\pm$ 0.001 & 0.135 $\pm$ 0.002 \\
		VDropout 				  & 96.2 $\pm$ 0.2 & 0.012 $\pm$ 0.001 & 0.128 $\pm$ 0.002 \\
		Rank-1 BNN				& 95.8 $\pm$ 0.1 & 0.011 $\pm$ 0.002 & 0.143 $\pm$ 0.004 \\
		\midrule
		SNGP 				& 95.9 $\pm$ 0.1 & 0.015 $\pm$ 0.001 & 0.143 $\pm$ 0.002 \\
		DUE 				& 95.6 $\pm$ 0.1 & 0.014 $\pm$ 0.006 & 0.170 $\pm$ 0.005  \\
		\midrule 
		\textit{Density Uncertainty} & & & \\
		Full-rank LDL	& 96.4 $\pm$ 0.1 & \textbf{0.010 $\pm$ 0.001} & 0.119 $\pm$ 0.003 \\
		Rank-1 Mixture ($1.25\%$)& \textbf{96.5 $\pm$ 0.0} & 0.011 $\pm$ 0.001 & \textbf{0.118 $\pm$ 0.001} \\
		\bottomrule
	\end{tabular}
\end{table*}

\begin{table*}[t]
	\caption{CIFAR-100 classification using the WRN28 architecture and 25 posterior samples. The average and the standard deviation over 3 random seeds are shown. Density Uncertainty improves all three metrics}
	\label{table:cifar100_wrn28}
	\centering
	\begin{tabular}{l c c c}
		\toprule
		Method 	& Accuracy ($\uparrow$)  & ECE ($\downarrow$) & NLL ($\downarrow$) \\
		\midrule
		MFVI 						 & 74.2 $\pm$ 0.8 & 0.260 $\pm$ 0.006 & 1.197 $\pm$ 0.023 \\
		MCDropout 				& 80.2 $\pm$ 0.3 & 0.033 $\pm$ 0.003 & 0.766 $\pm$ 0.005\\
		VDropout 				  & 81.1 $\pm$ 0.3 & 0.036 $\pm$ 0.002 & 0.756 $\pm$ 0.011 \\
		Rank-1 BNN				& 79.7 $\pm$ 0.2 & 0.032 $\pm$ 0.001 & 0.815 $\pm$ 0.008 \\
		\midrule
		SNGP 						 & 80.5 $\pm$ 0.3 & 0.035 $\pm$ 0.006 & 0.782 $\pm$ 0.017 \\
		DUE 				& \multicolumn{3}{c}{Out of memory error} \\
		\midrule 
		\textit{Density Uncertainty} & & & \\
		Full-rank LDL	  & \textbf{82.3 $\pm$ 0.2} & \textbf{0.029 $\pm$ 0.003} & \textbf{0.684 $\pm$ 0.005} \\
		Rank-1 Mixture ($1.25\%$) & \textbf{82.3 $\pm$ 0.0} & 0.032 $\pm$ 0.003 & 0.692 $\pm$ 0.001 \\
		\bottomrule
	\end{tabular}
\end{table*}

\subsection{CIFAR-10 and CIFAR-100}
For the CIFAR-10/100 experiments, we report accuracy, Expected Calibration Error (ECE) \citep{naeini2015obtaining} and Negative Log-Likelihood (NLL). 
ECE assesses the quality of the model's predictive uncertainty estimates by measuring how well-calibrated the model's predictions are. NLL is a proper scoring rule \citep{gneiting2007strictly} which favors predictions that are both accurate and well-calibrated.  
 
The results using ResNet-14 are summarized in \Cref{table:cifar10} and \Cref{table:cifar100}. On CIFAR-10, Density Uncertainty significantly reduces ECE compared to the baselines while maintaining comparable accuracy and slightly improving NLL. This indicates that Density Uncertainty provides more precise uncertainty estimates with lower calibration error. On CIFAR-100, Density Uncertainty improves the performance on metrics. Notably, Density Uncertainty again reduces ECE significantly. These results show that Density Uncertainty delivers the most reliable uncertainty estimates among the baselines. 

\Cref{table:cifar10_wrn28} and \Cref{table:cifar100_wrn28} present the results using the larger WRN-28 architecture, including the additional last layer GP baselines of SNGP \citep{liu2020simple} and DUE \citep{van2021feature}. For Density Uncertainty, we also experiment with the efficient rank-1 mixture Gaussians energy model to test the scalability of the method in addition to the default LDL parametrization. We set the number of mixture components to only $1.25\%$ of the layers' width, adding only negligible amount of parameter and computational overhead to the neural network. 
Density Uncertainty brings the best performance in all three metrics. Surprisingly, the rank-1 mixture with only $1.25\%$ number of components achieves performance comparable with the full-rank energy model, demonstrating the effectiveness of low-rank approximations in neural networks \citep{maddox2019simple,dusenberry2020efficient} and the scalability of the method. 
\Cref{table:rank1_ablation} shows the results on CIFAR-10 classification using the WRN28 architecture with varying number of rank-1 mixture components. We find that the performance is robust for a wide range of values, ranging from $1.25\%$ to $20\%$, and is comparable to the full-rank LDL model. 

\begin{table*}[t]
%	\small
	\caption{Varying the number of components for rank-1 Gaussian mixture energy on CIFAR-10 classification using the WRN28 architecture. The average and the standard deviation over 3 random seeds are shown. The performance of rank-1 Gaussian mixure is robust for a wide range of values}
	\label{table:rank1_ablation}
	\centering
	\begin{tabular}{l c c c}
		\toprule
		Method 	& Accuracy ($\uparrow$)  & ECE ($\downarrow$) & NLL ($\downarrow$) \\
		\midrule
		Rank-1 Mixture ($1.25\%$)	& 96.5 $\pm$ 0.0 & 0.011 $\pm$ 0.001 & 0.118 $\pm$ 0.001 \\
		Rank-1 Mixture ($2.5\%$)		 & 96.4 $\pm$ 0.1 & 0.012 $\pm$ 0.001 & 0.124 $\pm$ 0.002 \\
		Rank-1 Mixture ($5\%$)		  & 96.4 $\pm$ 0.1 & 0.012 $\pm$ 0.001 & 0.122 $\pm$ 0.001 \\
		Rank-1 Mixture ($10\%$)		& 96.5$\pm$ 0.0 & 0.012 $\pm$ 0.001 & 0.122 $\pm$ 0.001 \\
		Rank-1 Mixture ($20\%$)		& 96.5$\pm$ 0.0 & 0.013 $\pm$ 0.000 & 0.121 $\pm$ 0.003 \\
		\midrule
		Full-rank LDL	& 96.4 $\pm$ 0.1 & 0.010 $\pm$ 0.001 & 0.119 $\pm$ 0.003 \\
		\bottomrule
	\end{tabular}
\end{table*}

\begin{table*}[t]
	\caption{Out-of-distribution detection performance on SVHN using ResNet-14 models trained on CIFAR-10/100. The average and the standard deviation over 3 random seeds are shown. Using the energy statistics, Density Uncertainty gives the most robust out-of-distribution detection performance}
	\label{table:ood}
	\centering
	\begin{tabular}{l c c c c}
		\toprule
		& \multicolumn{2}{c}{CIFAR-10 $\rightarrow$ SVHN} & \multicolumn{2}{c}{CIFAR-100 $\rightarrow$ SVHN} \\
		\cmidrule(r){2-5}
		Method 	& AUPRC ($\uparrow$)  & AUROC ($\uparrow$) &  AUPRC ($\uparrow$)  & AUROC ($\uparrow$) \\
		\midrule
		MFVI 						 & 0.903 $\pm$ 0.009 & 0.830 $\pm$ 0.014 & 0.803 $\pm$ 0.008 & 0.640 $\pm$ 0.017\\
		MCDropout 				& 0.892 $\pm$ 0.006 & 0.832 $\pm$ 0.010 & 0.817 $\pm$ 0.005 & 0.666 $\pm$ 0.009 \\
		VDropout 				  & 0.917 $\pm$ 0.011 & 0.866 $\pm$ 0.018 & 0.822 $\pm$ 0.028 & 0.677 $\pm$ 0.052\\
		Rank-1 BNN				& 0.925 $\pm$ 0.003 & 0.880 $\pm$ 0.001 & 0.822 $\pm$ 0.027 & 0.681 $\pm$ 0.047\\
		\midrule 
		Density Uncertainty	  & \textbf{0.952 $\pm$ 0.026} & \textbf{0.893 $\pm$ 0.056} & \textbf{0.908 $\pm$ 0.015} & \textbf{0.800 $\pm$ 0.024} \\
		\bottomrule
	\end{tabular}
\end{table*}

\subsection{Out-of-Distribution Detection on SVHN} 
Can we deploy the energy models of the density uncertainty layers for detecting out-of-distribution (OOD) inputs?  
\Cref{table:ood} summarizes the the OOD detection performance on SVHN, using the models trained on CIFAR-10/100. We report the area under the precision-recall curve (AUPRC) and the receiver operator characteristic (AUROC). Following the previous work \citep{ritter2021sparse}, the baselines use the maximum predicted probability heuristic for OOD detection. On the other hand, Density Uncertainty, equipped with layer-wise generative energy models, can inherently perform OOD detection using the energy statistics. Specifically, the energy at layer $\ell$ can be represented as the squared sum of random variables:
\begin{align}
E_{\ell}(h_{\ell}) = \sum_{j=1}^D (z_\ell^j)^2 \text{ where } z_\ell = \Sigma_{\ell}^{-\frac{1}{2}} h_{\ell},
\end{align}
where $z_\ell$ can be thought of as whitened input with $\Sigma_{\ell}^{-1/2}$ decorrelating the layer's input $h_{\ell}$. Based on the NN-GP equivalence in the infinite width limit, we expect the energy to be approximately normally-distributed. Therefore, we use the squared deviation of energy from the in-distribution average
\begin{align}
| E_{\ell}(h_{\ell}) - \mu_{\ell} |^2
\end{align} 
as a test statistic for OOD detection where the in-distribution average energy $\mu_{\ell}$ is estimated using a in-distribution held-out set. We use the energy of the last convolutional layer as it reflects the most high-level, semantic information of the input. 
\Cref{table:ood} demonstrates that Density Uncertainty can most robustly detect OOD input among the baselines.

\section{Conclusion}
We proposed a novel density criterion for reliable uncertainty estimation, asserting that the predictive uncertainty of a model should be grounded in the empirical density of the input. A model that adheres to the criterion will produce higher uncertainty for inputs that are improbable in the training data, and lower uncertainty for those inputs that are more probable. 
We formalized the concept as a constraint on the predictive variance of a stochastic function and developed the density uncertainty layer as a flexible building block for uncertainty-aware deep learning. 
Through the empirical studies, we demonstrated that the proposed method provides the most reliable uncertainty estimates and robust out-of-distribution detection performance among the baselines. 
This could have practical applications in various fields where robust uncertainty estimation is crucial, such as medical diagnosis, autonomous driving, and financial forecasting.

%Looking ahead, there are several promising directions for future research. For example, exploring the use of mixture noise distributions with heavy tails \citep{dusenberry2020efficient} could enhance the model's robustness against data corruptions and distribution shift \citep{hendrycks2019benchmarking}. 
%In addition, incorporating density uncertainty layer into other deep neural network architectures such as Transformer \citep{vaswani2017attention} could prove effective for building uncertainty-aware models for sequence data such as natural language and times series. 

\bibliography{references}
\bibliographystyle{plainnat}

\appendix

%\Cref{sec:uci_regression} presents additional experiment results on the UCI regression benchmarks. 
%\Cref{sec:bnn} gives a brief overview of Bayesian Neural Networks and variational inference. 
%\Cref{sec:classification} studies Bayesian uncertainty in classification. 

\begin{table*}[t]
	\caption{Test NLL on UCI Regression. Lower is better. The average and the standard deviation over the 20 random train-test splits are shown}
	\label{table:uci_nll}
	\small
	\centering
	\begin{tabular}{l c c c c}
		\toprule
		Dataset 				 & MCDropout 				& VDropout 				& Rank1-BNN 			& Density Uncertainty \\
		\midrule
		Boston Housing 	    & 2.794 $\pm$ 0.141 & 2.815 $\pm$ 0.092 & 2.588 $\pm$ 0.236 & \textbf{2.523} $\pm$ 0.205 \\
		Concrete Strength  & 3.533 $\pm$ 0.037 & 3.497 $\pm$ 0.050 & 3.112 $\pm$ 0.081 & \textbf{3.093} $\pm$ 0.116 \\
		Energy Efficiency 	 & 2.604 $\pm$ 0.065 & 2.525 $\pm$ 0.057 & 2.044 $\pm$ 0.099 & \textbf{2.034} $\pm$ 0.087 \\
		Kin8nm 					& -0.590 $\pm$ 0.012 & -0.792 $\pm$  0.015 & -1.194 $\pm$ 0.028 & \textbf{-1.234} $\pm$ 0.036 \\
		Naval Propulsion   & -3.437 $\pm$ 0.048 & -4.019 $\pm$ 0.008 & -4.956 $\pm$ 0.033 & \textbf{-5.274} $\pm$ 0.036  \\
		Protein Structure   & 2.935 $\pm$ 0.006 & 2.852 $\pm$ 0.005 & \textbf{2.771} $\pm$ 0.010 & 2.821 $\pm$ 0.001 \\
		Wine Quality Red   & 0.952 $\pm$ 0.062 & \textbf{0.950} $\pm$ 0.067 & 0.954 $\pm$ 0.108 & 0.981 $\pm$ 0.109 \\
		Yacht Hydrodynamics & 3.205 $\pm$ 0.062 & 3.206 $\pm$ 0.071 & 2.712 $\pm$ 0.084 & \textbf{2.593} $\pm$ 0.067 \\
		Year Prediction MSD   & 3.568 $\pm$ NA & \textbf{3.558} $\pm$ NA & 3.566 $\pm$ NA & 3.570 $\pm$ NA \\
		\bottomrule
	\end{tabular}
\end{table*}

\begin{table*}[t]
	\caption{Test RMSE on UCI Regression. Lower is better. The average and the standard deviation over the 20 random train-test splits are shown}
	\label{table:uci_rmse}
	\small
	\centering
	\begin{tabular}{l c c c c}
		\toprule
		Dataset 				   & MCDropout 			& VDropout 				& Rank1-BNN 				& Density Uncertainty \\
		\midrule
		Boston Housing 		 & 3.715 $\pm$ 0.952 & 3.710 $\pm$ 0.729 & 3.212 $\pm$ 0.758 & \textbf{2.957} $\pm$ 0.606 \\
		Concrete Strength 	& 7.589 $\pm$ 0.630 & 7.154 $\pm$ 0.672 & 5.348 $\pm$ 0.552 & \textbf{5.290} $\pm$ 0.639 \\
		Energy Efficiency 	  & 3.073 $\pm$ 0.360 & 3.209 $\pm$ 0.429 & 1.718 $\pm$ 0.246 & \textbf{1.690} $\pm$ 0.268 \\
		Kin8nm 					 & 0.121 $\pm$ 0.003 & 0.094 $\pm$ 0.003 & 0.073 $\pm$ 0.003 & \textbf{0.070} $\pm$ 0.002 \\
		Naval Propulsion    & 0.007 $\pm$ 0.000 & 0.003 $\pm$ 0.000 & 0.010 $\pm$ 0.000 & \textbf{0.001} $\pm$ 0.000 \\
		Protein Structure    & 4.549 $\pm$ 0.033 & 4.123 $\pm$ 0.028 & \textbf{3.894} $\pm$ 0.046 & 4.077 $\pm$ 0.043 \\
		Wine Quality Red    & 0.626 $\pm$ 0.046 & 0.626 $\pm$ 0.049 & \textbf{0.623} $\pm$ 0.051 & 0.630 $\pm$ 0.050 \\
		Yacht Hydrodynamics & 5.060 $\pm$ 1.328 & 4.820 $\pm$ 1.175 & 3.070 $\pm$ 0.082 & \textbf{2.505} $\pm$ 0.060 \\
		Year Prediction MSD   & 8.726 $\pm$ NA & \textbf{8.700} $\pm$ NA & 8.712 $\pm$ NA & 8.710 $\pm$ NA \\
		\bottomrule
	\end{tabular}
\end{table*}

\begin{table*}[t]
	\caption{The size and the dimensionality of the UCI regression datasets}
	\label{table:uci_datasets}
	\small
	\centering
	\begin{tabular}{l c c }
		\toprule
		Dataset 				 & Size & Dimension \\
		\midrule
		Boston Housing 	    & 506 & 13 \\
		Concrete Strength  & 1,030 & 8 \\
		Energy Efficiency 	 & 768 & 8 \\
		Kin8nm 					& 8,192 & 8 \\
		Naval Propulsion   & 11,934 & 16 \\
		Protein Structure   & 45,730 & 9  \\
		Wine Quality Red   & 1,599 & 11 \\
		Yacht Hydrodynamics & 308 & 6 \\
		Year Prediction MSD   & 515,345 & 90 \\
		\bottomrule
	\end{tabular}
\end{table*}

\section{UCI Regression Results}
\label{sec:uci_regression}
We perform additional experiments on the UCI regression benchmarks \citep{dua2017uci}, one of the standard uncertainty estimation benchmarks for Bayesian neural networks. We follow the experiment protocol of \citet{gal2016dropout}. 
The benchmark includes regression datasets of varying size ranging from 300 to 515K, and input dimensions ranging from 4 to 90, as summarized in \Cref{table:uci_datasets}. 

\textbf{Experiment details} 
We follow the experiment protocol of \citet{gal2016dropout}. We report the negative log-likelihood (NLL) and the root mean squared error (RMSE) on the test split using 10 posterior samples. 
For each dataset, the results are averaged over 20 random train-test splits of the data (except for Protein which uses 5 splits and Year which uses a single split). 
We normalize the input using the mean and the standard deviation of the training split. 
We use MLPs with two hidden layers of width 50. We increase the width to 100 for the larger datasets of Protein and Year. 
All models are trained for 100 epochs with learning rate of 0.01 using momentum optimizer. We use batch size of 128 and weight decay of 0.0001. 
We treat the output variance as a trainable parameter. 
In order to minimize hyperparameter tuning, we use the same hyperparameters used in the CIFAR-10/100 experiments except that we initialize the posterior standard deviations to a higher value of 0.1 as lower values led to overfitting. 

\textbf{Results} \Cref{table:uci_nll} and \Cref{table:uci_rmse} present the test NLL and the test RMSE on the datasets. Overall, Density Uncertainty gives the best results, outperforming the baselines on 6 of 9 datasets in both NLL and RMSE. 
These results demonstrate that density uncertainty layers can also bring benefits for regression problems potentially in low-data regimes besides the natural image classification problems studied in the paper. 
However, a caveat is that the variance of results are high on some datasets due to their small sizes and the performance on these datasets can be sensitive to the hyperparameters.

\section{Proof of Proposition 1}
For brevity, we omit the noise variable $\epsilon_\ell$ and the parameter $\phi_\ell$ from the layer $f_\ell(h_\ell; \epsilon_\ell; \phi_\ell)$.
First decompose the network output variance as
\begin{align}
&\text{Var}[a_{L+1}^j|x] \\
&= \text{Var}[a_{L}^j + f_L^j(h_L) | x] \\
&= \text{Var}[a_{L}^j|x] + \text{Var}[f_L^j(h_L)| x] + 2 \text{Cov}[a_L^j, f_L^j(h_L) | x] \nonumber 
\end{align}
The second term can be decomposed as
\begin{align}
&\text{Var}[f_L^j(h_L)| x] \\ 
&= \mathbb{E}[\text{Var}[f_L^j(h_L)|h_L]] + \text{Var}[\mathbb{E}[f_L^j(h_L)|h_L]].
\end{align}
Each of these components are bounded as
\begin{align}
\alpha \mathbb{E}[E_{L}(h_L)] \le \mathbb{E}[\text{Var}[f_L^j(h_L)|h_L]] \le \beta \mathbb{E}[E_{L}(h_L)], \nonumber
\end{align}
by assumption, and 
\begin{align}
0 \le \text{Var}[\mathbb{E}[f_L^j(h_L)|h_L]] \le MD \max_k \text{Var}[a_{L}^k|x]
\end{align}
%And the second term is
%\begin{align}
%&\text{Var}[\mathbb{E}[f_L^j(h_L)|h_L]] \le (w_L^j)^T \text{Cov}[a_L] w_L^J
%\end{align}

On the other hand, the covariance term is bounded as
\begin{align}
\Big|\text{Cov}[a_L^j, f_L^j(h_L) | x]\Big| \le \sqrt{M} D \max_k \text{Var}[a_L^k|x]
%&= (w_L^j)^T \text{Cov}[a_L^j, \phi(a_L)|x)] \\
%&\ge  -\|w_L^j\|_2 \|\text{Cov}[a_L^j, a_L]\|_2 \\
\end{align}
If there exist $C_L, C_L'$ such that for all $j$,
\begin{align}
C_L \mathbb{E}[E(x_1, ..., h_{L-1})] \le \text{Var}[a_L^j|x] 
\end{align}
and
\begin{align}
\text{Var}[a_L^j|x] \le C_L' \, \mathbb{E}[E(x_1, ..., h_{L-1})].
\end{align}
Combining the above, we have 
\begin{align}
\text{Var}[a_{L+1}^j|x] \ge C \, \mathbb{E}[E(x_1, ..., h_{L})],
\end{align}
for some constant $C$ by induction. 

\section{Bayesian Uncertainty in Linear Classification}
\label{sec:classification}
In the paper, we demonstrated that the Bayesian uncertainty for linear regression is grounded in the density estimate of input. But how about in classification?
We show that the Bayesian uncertainty in classification is also based on a density estimate but with a weighted generative objective. 

Consider a logistic classification problem, with the input $\X \in \mathbb{R}^{N \times D}$, the target $\y \in \mathbb{R}^N$, and the weight $w \in \mathbb{R}^D$:
\begin{align}
p(w) 				&= \mathcal{N}(w| 0, \alpha^{-1} I), \label{eq:bayesian_classification_weight} \\ 
p(\y | \X, w)  &= \prod_{i=1}^N y_i^{\sigma(w^T x_i)}(1 - y_i)^{1 - \sigma(w^T x_i)}, \label{eq:bayesian_classification_output} 
\end{align}
where $\sigma$ is the logistic sigmoid function: $\sigma(x) = 1 / (1 + e^{-x})$. Although the exact posterior distribution is intractable in this case, we can obtain a Gaussian approximation using the Laplace's approximation \citep{bishop06}:
\begin{gather}
q(w) = \mathcal{N}(w | \mu_{\text{MAP}}, \Lambda^{-1}), \\
\Lambda = \sum_{i=1}^N \sigma(w^T x_i) (1 - \sigma(w^T x_i)) x_i x_i^T + \alpha I, \label{eq:classification_lambda}
\end{gather}
and $\mu_{\text{MAP}}$ is the MAP estimate of the weight. 
Comparing the posterior precision (\Cref{eq:classification_lambda}) to that of the regression in the paper $\Lambda = \beta \sum_{i} x_i x_i^T + \alpha I$, we find that the precision matrix in classification is a \textit{weighted} estimate of the input covariance. 
Noting that the weight $\sigma(w^T x_i) (1 - \sigma(w^T x_i))$ is higher for inputs with more uncertain predictions (i.e., $\sigma(w^T x_i)$ is closer to $0.5$), Bayesian classification prioritizes inputs that give higher prediction uncertainty, in contrast to the regression case where all inputs were weighted uniformly. 
Recall that the posterior precision essentially serves as a Gaussian density estimate of the input. \Cref{eq:classification_lambda} shows that Bayesian logistic classification performs weighted generative modeling of the input, and by prioritizing inputs that are more informative, it can potentially better utilize the capacity of the Gaussian energy model. 

Despite this finding, in this work we choose to use the unweighted generative objective for density uncertainty layers because (1) the weighted objective can lead to an unfaithful estimate of the input density and (2) the capacity of the generative model is not a major concern for density uncertainty layers, as the complexity of the uncertainty landscape naturally grows with the number of layers in the network.  

%\bibliography{references}
%\bibliographystyle{plainnat}

%\input{checklist}

\end{document}